\documentclass[11pt,a4paper,nonumbering]{ds}

\usepackage[authoryear,sort&compress,round]{natbib}
\usepackage{graphicx}
\usepackage{booktabs}
\usepackage{multirow}
\usepackage{array}
\usepackage{tabularx}
\usepackage{makecell}
\usepackage{float}
\usepackage{dblfloatfix}
\usepackage{caption}
\usepackage{xcolor}
\usepackage{enumitem}
\usepackage{hyperref}

\renewcommand{\today}{}
\reportnumber{001}

\newcommand{\echoname}{Echo-$\alpha$}
\newcommand{\echoground}{Echo-$\alpha$-Grounding}
\newcommand{\echodiag}{Echo-$\alpha$-Diagnosis}
\newcommand{\specialdet}{Specialized Detector}
\newcommand{\directtool}{Direct MLLM + tool}

\title{\centering \echoname: Large Agentic Multimodal Reasoning Model for Ultrasound Interpretation}

\makeatletter
\def\@author{
\begin{center}
\normalsize
Jing Zhang$^{1,*}$, Wentao Jiang$^{1,*}$, Tao Huang$^1$, Zhiwei Wang$^1$, Jianxin Liu$^2$, Jian Chen$^2$, Ping Ye$^2$,
Gang Wang$^3$, Zengmao Wang$^1$, Bo Du$^1$, Dacheng Tao$^4$
\\[4pt]
{\footnotesize
\renewcommand{\arraystretch}{1.0}
\begin{tabular}{c}
$^1$School of Computer Science, Wuhan University, China\\
$^2$The Central Hospital of Wuhan, China\\
$^3$Taizhou Hospital of Zhejiang Province, China\\
$^4$Nanyang Technological University, Singapore
\end{tabular}}
\\[4pt]
{\small \texttt{jingzhang.cv@gmail.com, jiang\_wentao@whu.edu.cn}}
\end{center}
}
\makeatother

\correspondingauthor={*These authors contributed equally to this work.}
\fancypagestyle{firststyle}{
    \fancyhead[R]{}
    \fancyhead[L]{{\footerfont\itshape\monthyeardate\today}}
    \fancyhead[C]{}
    \fancyfoot[L]{\footerfont *These authors contributed equally to this work.}
    \fancyfoot[R]{}
    \fancyfoot[C]{}
}

\begin{abstract}
Ultrasound interpretation requires both precise lesion localization and holistic clinical reasoning, yet existing methods typically excel at only one of these capabilities: specialized detectors offer strong localization but limited reasoning, whereas multimodal large language models (MLLMs) provide flexible reasoning but weak grounding in specialized medical domains.
We present \textbf{\echoname}, an agentic multimodal reasoning model for ultrasound interpretation that unifies these strengths within an invoke-and-reason framework.
\echoname{} is trained to coordinate organ-specific detector outputs, integrate them with global visual context, and convert the resulting evidence into grounded diagnostic decisions beyond detector-only inference.
This behavior is established through a nine-task supervised curriculum and then refined by sequential reinforcement learning under different reward trade-offs, yielding \textbf{\echoground} for lesion anchoring and \textbf{\echodiag} for final diagnosis.
On multi-center renal and breast ultrasound benchmarks, \echoname{} outperforms competitive baselines on both grounding and diagnosis.
In particular, on cross-center test sets, \echoground{} attains 56.73\%/43.78\% F1@0.5 and \echodiag{} reaches 74.90\%/49.20\% overall accuracy on renal/breast ultrasound.
These results suggest that agentic multimodal reasoning can turn specialized detectors into verifiable clinical evidence, offering a practical route toward ultrasound AI systems that are more accurate, interpretable, and transferable.
The repository is at \url{https://github.com/MiliLab/Echo-Alpha}.
\end{abstract}

\begin{document}
\maketitle

\section{Introduction}
Ultrasound is one of the most widely used diagnostic modalities in clinical medicine, valued for its non-invasiveness, portability, and cost-effectiveness, but its interpretation remains highly operator-dependent and requires extensive clinical expertise~\citep{survey_ultrasound}.
These factors have motivated substantial efforts on automated ultrasound analysis, from early computer-aided detection (CAD) systems~\citep{survey_cad} to modern deep-learning models~\citep{utnet,sam_med,surgicalsam,unet++}.

Despite impressive progress, existing approaches face two fundamental limitations.
Specialized lesion detection models trained on specific anatomical regions provide strong localization performance, but offer limited support for clinical reasoning, interpretability, and holistic assessment.
In contrast, recent multimodal large language models (MLLMs)~\citep{llava-med,llm-med} exhibit promising visual understanding and reasoning capabilities, yet remain weak at fine-grained spatial grounding in medical images and prone to hallucination in specialized domains~\citep{hallu-framework,med-halt}.

A natural way to bridge these complementary weaknesses is the agentic paradigm, in which a general-purpose reasoning model coordinates domain-specific tools~\citep{toolformer,radagents}.
Recent work~\citep{Deepseek-R1,med_r1,pearl} further suggests that reinforcement learning can substantially improve reasoning and tool-use behavior.
However, it remains underexplored how to close the loop between tool invocation and spatial grounding, where the model must decide when to call a detector, how to verify its outputs, and how to integrate detector evidence with broader visual context to form grounded diagnoses.

We present \echoname, an agentic multimodal reasoning model for ultrasound interpretation.
At its core, \echoname{} places the MLLM in the role of a central reasoning module within an invoke-and-reason loop: conditioned on a raw ultrasound image together with optional clinical context such as patient history and indication, it coordinates organ-specific detection tools, integrates their outputs with global visual context, and produces grounded clinical decisions that are not limited to detector-only inference.
This formulation is motivated by the nature of ultrasound interpretation, where reliable diagnosis depends not only on precise lesion localization but also on broader organ context and coupled sonographic cues.
To reflect this structure, training is organized in two stages.
First, a nine-task supervised curriculum equips the model with capabilities spanning spatial grounding, diagnostic reasoning, detector collaboration, report generation, and agentic tool interaction.
Starting from this shared initialization, reinforcement learning is then applied with different reward trade-offs, yielding \echoground{} for lesion anchoring and \echodiag{} for final diagnosis.

We validate \echoname{} on renal and breast ultrasound benchmarks under a multi-center protocol, where validation is performed in-center and testing is performed cross-center.
Across both organs, \echoground{} improves lesion localization over direct prompting baselines and strong specialized detectors, while the final diagnosis model \echodiag{} improves overall category prediction by integrating grounded evidence with global visual context.
Taken together, these results suggest that agentic multimodal reasoning can transform specialized detectors from fixed predictors into verifiable clinical evidence.
By first anchoring lesion evidence and then optimizing for clinical decision-making, \echoname{} yields more reliable and interpretable diagnoses than either detector-only or generic MLLM baselines.

\section{Related Work}
\subsection{Tool-Augmented MLLMs}
Tool augmentation provides a natural way to combine the flexibility of large models with structured external evidence.
Toolformer~\citep{toolformer} first showed that language models can learn when and how to invoke external tools during generation, establishing a general paradigm for tool-aware reasoning.
In medical imaging, ChatCAD and ChatCAD+~\citep{chatcad,chatcad+} demonstrated that interactive diagnosis can benefit from coupling a language model with specialized medical modules, enabling multi-turn consultation and more structured decision support.
RadAgents~\citep{radagents} further organizes multimodal radiology reasoning into a workflow with explicit intermediate steps and external components, highlighting the value of agentic orchestration in clinical imaging.
Our work follows this line and studies tool use in ultrasound, where detector feedback must be interpreted and refined before producing the final grounded diagnosis.

\subsection{Medical MLLMs}
Medical MLLMs have advanced rapidly in both multimodal understanding and medical reasoning.
LLaVA-Med~\citep{llava-med} adapts the LLaVA framework to the biomedical domain and shows that instruction tuning can substantially improve medical visual-language interaction, while BiomedCLIP~\citep{biomedclip} learns large-scale biomedical image-text alignment and provides strong cross-modal representations for downstream medical tasks.
In the ultrasound domain, Sonomate~\citep{sonomate} explores visually grounded language modeling for fetal ultrasound understanding, and EchoVLM~\citep{echovlm} studies ultrasound-specific multimodal modeling with a mixture-of-experts design.
Beyond representation learning, RL-enhanced medical reasoning models such as Med-R1~\citep{med_r1} indicate that reward-based optimization can further strengthen multi-step medical decision-making.
Building on these developments, \echoname{} emphasizes grounded ultrasound diagnosis under an agentic tool-use framework.

\section{Method}
\subsection{Overview}
\echoname{} is built on a pretrained MLLM and operates as an ultrasound diagnostic agent equipped with domain-specific tools.
Its core workflow follows an invoke-and-reason loop.
Given a raw ultrasound image together with optional clinical context such as patient history, the model first performs visual reasoning to form an initial hypothesis, then invokes a specialized detection tool via a structured function call to obtain candidate bounding boxes and category predictions.
The tool returns (i) a rendered visualization with overlaid detections and (ii) structured metadata including coordinates, confidence scores, and labels.
The model then compares the tool feedback against its own perception and produces a grounded output.

\begin{figure}[t]
\centering
\includegraphics[width=\linewidth]{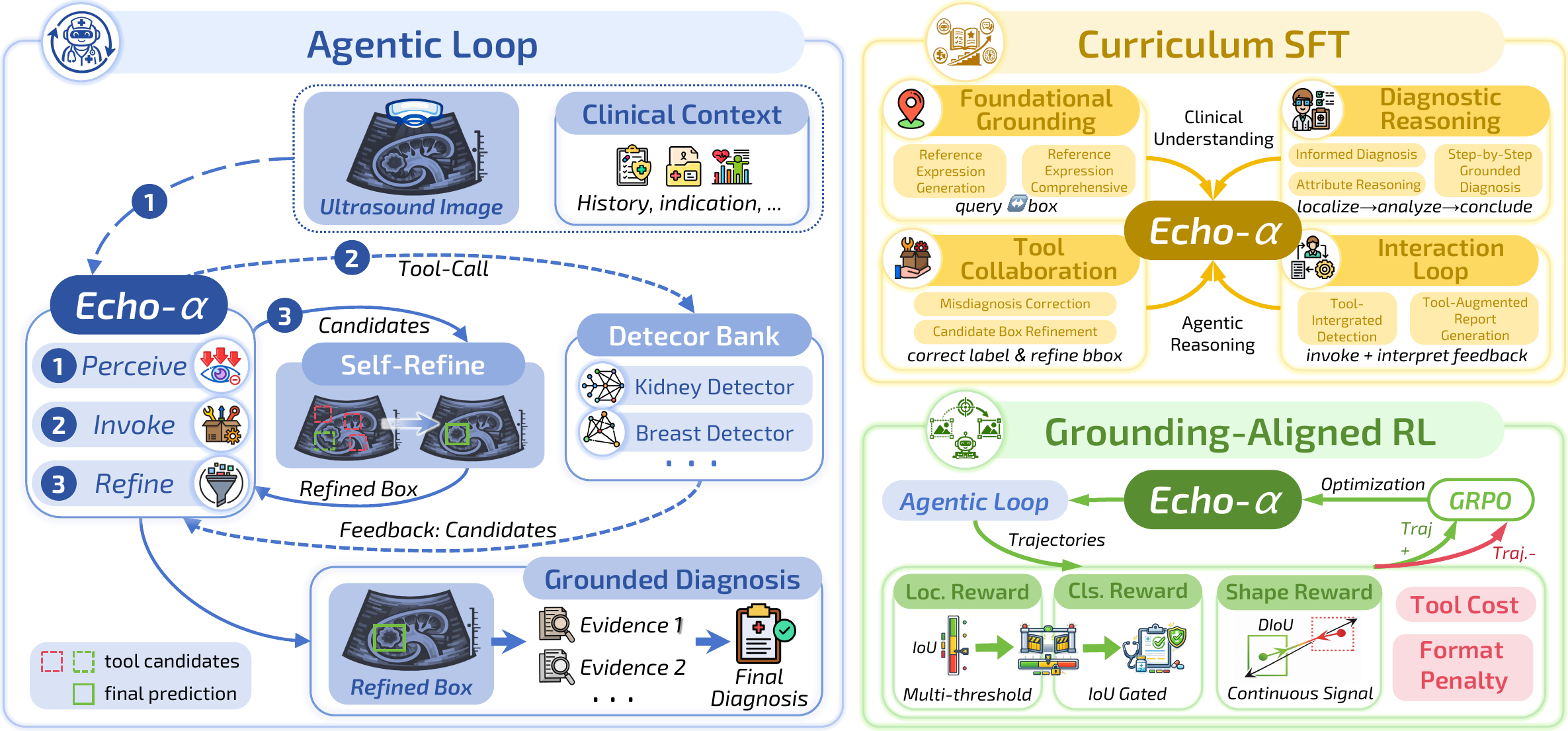}
\caption{\textbf{Overview of \echoname.} \echoname{} follows an invoke-and-reason loop that alternates between internal visual reasoning and calls to organ-specific detectors. The supervised curriculum covers foundational grounding, diagnostic reasoning, tool collaboration and report generation, enabling the model to interpret detector candidates, refine boxes, correct labels, and produce grounded diagnoses. Reinforcement learning then further aligns the full interaction trajectory with localization quality, category correctness, geometric consistency, and efficient tool use.}
\label{fig:overview}
\end{figure}

Figure~\ref{fig:overview} shows the unified \echoname{} framework.
At a high level, the model first forms an initial hypothesis from the image, then consults the detector bank, interprets the returned candidates, and finally synthesizes a grounded diagnostic output.
The SFT stage teaches these skills explicitly through a multi-task curriculum, while the RL stage optimizes the complete interaction loop under grounding- and diagnosis-related rewards.

Within this unified framework, the difference between grounding and diagnosis is introduced only through the RL objective.
Both settings share the same SFT model as the starting point, and are then optimized by shifting how the reward weights balance lesion grounding against final diagnosis.
This design is motivated by the nature of ultrasound interpretation.
The model must first anchor the lesion region reliably, yet the final diagnosis cannot be determined from the local box alone and instead depends on clinical context, broader organ-level information, and multiple coupled sonographic cues.
Because these goals are difficult to optimize equally well at the same time, the model is first optimized toward grounding and then toward the final diagnosis.

\subsection{Lesion Detection Tools}
For each anatomical domain, we deploy a strong specialized lesion detector as a callable tool.
Concretely, our detectors are built on LW-DETR~\citep{lw-detr} and further strengthened with improved model design, architecture choices, and training recipe, rather than using the original off-the-shelf configuration.
In the renal setting, the detector covers six lesion categories (Angiomyolipoma, Hydronephrosis, Renal Stone, Renal Cyst, Diffuse Renal Parenchymal Disease, and Renal Malignant Tumor).
In the breast setting, the detector outputs BI-RADS assessments across six categories (BI-RADS 2--5, with BI-RADS 4 further resolved into 4a/4b/4c).
All detectors expose a unified function-calling interface, making it straightforward to extend \echoname{} to additional organs by swapping in new tools without modifying the core agent.

\subsection{Stage 1: Supervised Fine-Tuning}
The supervised fine-tuning (SFT) stage equips \echoname{} with complementary skills through a nine-task curriculum organized into four tiers.

\textbf{Foundational grounding} includes Referring Expression Comprehension (REC) and Referring Expression Generation (REG)~\citep{mao2016generation}, establishing basic spatial grounding: given a textual query, predict the bounding box of the described lesion; conversely, given a bounding box, describe and classify its content.
\textbf{Diagnostic reasoning} progresses from direct diagnosis to attribute-based explanations and multi-step grounded analysis.
\textbf{Tool collaboration} trains the model to utilize detector outputs: refining imprecise boxes, correcting misclassified findings, and jointly assessing localization and classification.
\textbf{Interaction loop} teaches the model to actively invoke the tool, interpret the returned visualization, and synthesize a final grounded diagnosis.

Training data is constructed by prompting a large teacher model with ground-truth annotations and specialized-detector predictions as teacher-forcing context; the teacher-generated rationales are used as supervised targets.
After SFT, we obtain a single shared initialization that serves as the starting point for all subsequent RL training.

\subsection{Stage 2: Reinforcement Learning}
Starting from the same SFT-initialized model, the reinforcement learning (RL) stage trains the model to execute the complete agentic loop autonomously, using Group Relative Policy Optimization (GRPO)~\citep{grpo}.

The reward comprises three components.
The \textbf{localization reward} is a smoothed function of the intersection-over-union (IoU) between the predicted and ground-truth boxes, with a non-zero base score once IoU exceeds a small threshold.
This encourages any meaningful spatial overlap while providing a smooth gradient toward higher-quality localization.
The \textbf{classification reward} is activated only when the prediction is sufficiently localized and the category is correct, aligning label accuracy with spatial precision.
The \textbf{shape reward} uses Distance-IoU (DIoU)~\citep{diou} to provide a tighter and more continuous signal, encouraging compact alignment and mitigating sparse rewards.
Finally, each tool invocation incurs a small fixed cost, discouraging redundant calls while preserving the incentive to use the detector strategically.
We use the same reward components for both settings, but assign them different weights.

When the objective places more emphasis on localization- and shape-related rewards, the resulting model is better suited for lesion anchoring and box refinement; we refer to this model as \echoground.
Starting from this grounded model, we further shift the reward emphasis toward diagnosis-related correctness while preserving the localization terms, producing the final \echodiag{} model.

During RL training, we apply geometric augmentation for robustness, including random horizontal flipping, random resizing, random cropping, and square resizing.
These augmentations are used only in the RL stage.

\section{Experiments}
\subsection{Datasets and Metrics}
We evaluate \echoname{} on two ultrasound benchmarks.
\textbf{Renal Ultrasound} is annotated in COCO format with six lesion categories.
\textbf{Breast Ultrasound} is annotated with six BI-RADS categories.
For both benchmarks, we adopt a multi-center protocol.
The Val split comes from the same center as Train and measures in-center generalization, while the Test split comes from another center and measures cross-center robustness.
All models are trained on the training split only.

Each image contains either annotated lesions of interest (positive) or no lesion (negative).
For positive images, a prediction is correct if its IoU with the ground-truth box exceeds a threshold and the category matches.
For negative images, an empty detector output or an MLLM prediction of the negative class is counted as correct.
Detector-based diagnosis uses the top-ranked prediction, while MLLM-based grounding outputs a single box.
For the \textbf{Grounding} task, we report instance-level F1 aggregated over positive lesion instances and image-level accuracy over all images at IoU thresholds 0.25/0.5/0.75.
For the \textbf{Diagnosis} task, we report overall accuracy over all diagnosis classes, including the negative class, in the main paper.

\subsection{Baselines}
We compare the following configurations:
(1) \textbf{\specialdet} -- our strong organ-specific detector, built on LW-DETR~\citep{lw-detr} with improved design, architecture, and training recipe;
(2) \textbf{Direct MLLM} -- the backbone MLLM prompted directly without tool access or fine-tuning;
(3) \textbf{Direct MLLM with tool} -- the same model given tool access without fine-tuning;
(4) \textbf{SFT} -- the shared SFT model without RL adaptation;
(5) \textbf{SFT with tool} -- the SFT model with tool access at inference;
(6) \textbf{SFT+RL with tool (\echoname)} -- the final RL-optimized model.
For the generic MLLM baselines, we adopt Qwen3-VL~\citep{Qwen3-VL} as the backbone.
To ensure a fair comparison, all MLLM-based baselines use the same prompt template, tool-call setting, and output schema for grounded predictions.

\begin{table}[t]
\centering
\caption{\textbf{Grounding results on Renal and Breast Ultrasound.} `Val' denotes same-center validation and `Test' denotes cross-center testing. `F1@.25/.5/.75' and `Acc@.25/.5/.75' denote instance-level F1 and image-level accuracy at IoU thresholds 0.25, 0.5, and 0.75, respectively. Our final grounding model is reported as \echoground, optimized by RL.}
\label{tab:grounding}
\setlength{\tabcolsep}{1.5pt}
\renewcommand{\arraystretch}{1.08}
\begin{tabular}{@{}p{0.49\linewidth}@{\hspace{0.01\linewidth}}p{0.49\linewidth}@{}}
\begin{minipage}{\linewidth}
\centering
{\small\bfseries (a) Renal Ultrasound, Val}\\[2pt]
{\fontsize{7}{8}\selectfont
\begin{tabular}{@{}lcccccc@{}}
\toprule
Method & F1@.25 & F1@.5 & F1@.75 & Acc@.25 & Acc@.5 & Acc@.75\\
\midrule
\specialdet & 71.83 & 69.70 & 43.20 & 72.11 & 70.43 & 49.53\\
Direct MLLM & 18.93 & 21.71 & 16.48 & 17.32 & 16.67 & 12.85\\
\directtool & 59.33 & 56.56 & 36.34 & 60.45 & 58.30 & 42.63\\
SFT & 63.76 & 54.81 & 22.44 & 71.64 & 64.83 & 40.21\\
SFT + tool & 72.56 & 69.53 & 40.46 & 71.46 & 69.12 & 46.74\\
\echoground & \textbf{73.38} & \textbf{70.78} & \textbf{45.05} & \textbf{73.23} & \textbf{71.18} & \textbf{50.93}\\
\bottomrule
\end{tabular}
}
\end{minipage}
&
\begin{minipage}{\linewidth}
\centering
{\small\bfseries (b) Renal Ultrasound, Test}\\[2pt]
{\fontsize{7}{8}\selectfont
\begin{tabular}{@{}lcccccc@{}}
\toprule
Method & F1@.25 & F1@.5 & F1@.75 & Acc@.25 & Acc@.5 & Acc@.75\\
\midrule
\specialdet & 64.75 & 52.63 & 27.91 & 66.04 & 56.11 & 35.30\\
Direct MLLM & 15.66 & 10.10 & 1.44 & 14.05 & 9.60 & 1.38\\
\directtool & 61.49 & 50.11 & 24.69 & 63.40 & 52.66 & 33.97\\
SFT & 47.61 & 36.30 & 10.73 & 43.49 & 33.15 & 9.80\\
SFT + tool & 65.45 & 53.26 & 28.88 & 65.91 & 57.85 & 35.03\\
\echoground & \textbf{69.00} & \textbf{56.73} & \textbf{30.76} & \textbf{69.80} & \textbf{59.60} & \textbf{37.72}\\
\bottomrule
\end{tabular}
}
\end{minipage}
\end{tabular}
\par\vspace{7pt}
\begin{tabular}{@{}p{0.49\linewidth}@{\hspace{0.01\linewidth}}p{0.49\linewidth}@{}}
\begin{minipage}{\linewidth}
\centering
{\small\bfseries (c) Breast Ultrasound, Val}\\[2pt]
{\fontsize{7}{8}\selectfont
\begin{tabular}{@{}lcccccc@{}}
\toprule
Method & F1@.25 & F1@.5 & F1@.75 & Acc@.25 & Acc@.5 & Acc@.75\\
\midrule
\specialdet & 49.32 & 46.68 & \textbf{27.88} & 50.25 & 47.84 & \textbf{30.65}\\
Direct MLLM & 7.24 & 1.81 & 0.00 & 9.39 & 7.82 & 7.61\\
\directtool & 39.90 & 36.61 & 19.95 & 38.87 & 36.21 & 22.84\\
SFT & 15.56 & 11.01 & 0.97 & 15.32 & 10.99 & 9.41\\
SFT + tool & 43.59 & 40.64 & 23.85 & 44.52 & 41.86 & 26.99\\
\echoground & \textbf{52.96} & \textbf{50.37} & 25.79 & \textbf{54.73} & \textbf{52.49} & 30.40\\
\bottomrule
\end{tabular}
}
\end{minipage}
&
\begin{minipage}{\linewidth}
\centering
{\small\bfseries (d) Breast Ultrasound, Test}\\[2pt]
{\fontsize{7}{8}\selectfont
\begin{tabular}{@{}lcccccc@{}}
\toprule
Method & F1@.25 & F1@.5 & F1@.75 & Acc@.25 & Acc@.5 & Acc@.75\\
\midrule
\specialdet & 43.30 & 42.01 & \textbf{29.07} & 45.23 & 44.06 & \textbf{32.53}\\
Direct MLLM & 14.29 & 9.81 & 1.05 & 11.88 & 8.70 & 2.43\\
\directtool & 38.46 & 37.25 & 24.33 & 40.75 & 39.64 & 28.66\\
SFT & 22.13 & 20.98 & 13.45 & 24.86 & 23.76 & 16.85\\
SFT + tool & 41.78 & 40.11 & 24.99 & 43.99 & 42.54 & 29.42\\
\echoground & \textbf{45.72} & \textbf{43.78} & 28.37 & \textbf{47.55} & \textbf{45.32} & 31.70\\
\bottomrule
\end{tabular}
}
\end{minipage}
\end{tabular}
\end{table}

\subsection{Main Results}
Tables~\ref{tab:grounding} and~\ref{tab:diagnosis} summarize results for the two tasks under the multi-center evaluation protocol.
\echoname{} achieves the best F1@0.5 on all four grounding splits, reaching 70.78\%(+14.22), 56.73\%(+6.62), 50.37\%(+13.76), and 43.78\%(+6.53), compared with \directtool.
For diagnosis, the final model \echodiag{} consistently improves over direct MLLM baselines and reaches 77.43\%(+13.44), 74.90\%(+7.91), 48.75\%(+11.04), and 49.20\%(+4.45) overall accuracy on renal Val, renal Test, breast Val, and breast Test, respectively, compared with \directtool.
Overall, these results show that the proposed agentic framework improves both grounded localization and final diagnosis under the reported evaluation setting.
We summarize the main insights from these comparisons.

\begin{itemize}[leftmargin=1.2em]
\item \textbf{Direct prompting is insufficient.} Direct MLLM performs poorly on both grounding and diagnosis, showing that ultrasound understanding requires explicit adaptation rather than generic multimodal reasoning alone.
\item \textbf{Tool access helps, but learned tool use helps more.} \directtool{} already improves substantially over Direct MLLM, while SFT + tool further improves across all grounding splits, indicating that the model must learn how to interpret and refine detector feedback.
\item \textbf{RL brings consistent gains on top of tool-aware SFT.} Across all settings, RL further improves grounding performance over the tool-aware SFT model, indicating that reward-driven optimization helps the agent refine detector outputs beyond supervised imitation alone.
\item \textbf{Grounded evidence supports final diagnosis.} \echodiag{} consistently improves over direct MLLM diagnosis baselines, supporting the view that final ultrasound diagnosis requires not only lesion anchoring but also broader contextual reasoning.
\item \textbf{The gains transfer across centers.} Improvements are preserved on the cross-center test sets for both organs, suggesting that the model learns a transferable way to combine detector feedback with multimodal reasoning rather than merely fitting a single-center distribution.
\end{itemize}

\begin{table}[t]
\centering
\caption{\textbf{Diagnosis results on Renal and Breast Ultrasound.} Left: overall accuracy (\%) on the four evaluation splits, where `Val' denotes same-center validation and `Test' denotes cross-center testing. `Spec.', `Direct', and `Direct+Tool' denote Specialized Detector, Direct MLLM, and Direct MLLM with tool, respectively. Right: compact class-wise diagnosis trends for the same results.}
\label{tab:diagnosis}
\setlength{\tabcolsep}{4pt}
\begin{tabular}{@{}m{0.32\linewidth}m{0.63\linewidth}@{}}
\begin{minipage}[c]{\linewidth}
\centering
\vspace*{10pt}
\small
\renewcommand{\arraystretch}{1.05}
\begin{tabular}{@{}lr@{}}
\multicolumn{2}{c}{\bfseries (a) Renal Ultrasound, Val}\\
\toprule
Method & Overall\\
\midrule
Spec. & 74.53\\
Direct & 18.69\\
Direct+Tool & 63.99\\
\echodiag & \textbf{77.43}\\
\bottomrule
\end{tabular}

\vspace{7pt}
\begin{tabular}{@{}lr@{}}
\multicolumn{2}{c}{\bfseries (b) Renal Ultrasound, Test}\\
\toprule
Method & Overall\\
\midrule
Spec. & 69.13\\
Direct & 15.84\\
Direct+Tool & 66.99\\
\echodiag & \textbf{74.90}\\
\bottomrule
\end{tabular}

\vspace{7pt}
\begin{tabular}{@{}lr@{}}
\multicolumn{2}{c}{\bfseries (c) Breast Ultrasound, Val}\\
\toprule
Method & Overall\\
\midrule
Spec. & \textbf{51.41}\\
Direct & 14.44\\
Direct+Tool & 37.71\\
\echodiag & 48.75\\
\bottomrule
\end{tabular}

\vspace{7pt}
\begin{tabular}{@{}lr@{}}
\multicolumn{2}{c}{\bfseries (d) Breast Ultrasound, Test}\\
\toprule
Method & Overall\\
\midrule
Spec. & 46.96\\
Direct & 28.87\\
Direct+Tool & 44.75\\
\echodiag & \textbf{49.20}\\
\bottomrule
\end{tabular}
\end{minipage}
&
\begin{minipage}[c]{\linewidth}
\centering
\raisebox{0.01\height}{\hspace*{0.005\linewidth}\includegraphics[width=0.85\linewidth]{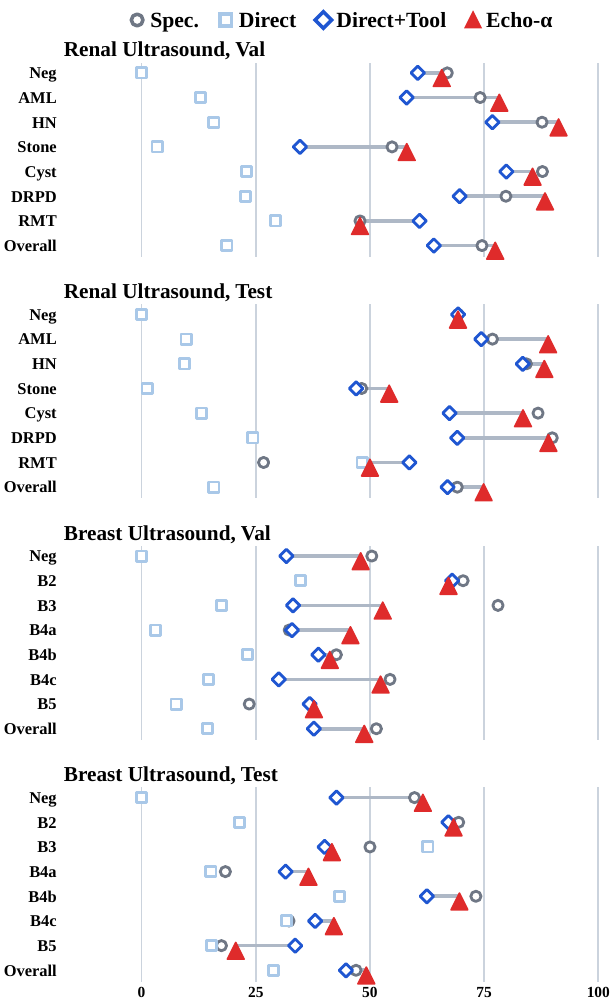}}
\end{minipage}
\end{tabular}
\end{table}

\subsection{Detector-Agnostic Tool Generalization}
To examine whether the learned invoke-and-reason behavior depends on a single specialized detector, we replace the renal detector tool with several detector backbones, including YOLO-family detectors~\citep{yolo,yolov11,yolov12,yolo26}, LW-DETR-DINOv3~\citep{lw-detr,dinov3}, and RF-DETR-DINOv3~\citep{rf-detr,dinov3}, and evaluate diagnosis accuracy under the same renal validation protocol.
Table~\ref{tab:detector_ablation} shows a clear and consistent benefit from \echoname{}: for every evaluated detector, adding the agentic reasoning module improves diagnosis accuracy, with gains ranging from +2.15 to +22.95 percentage points.
This consistent positive shift is the central observation of this ablation.
It indicates that \echoname{} does not simply inherit the detector's prediction, but turns detector outputs into revisable evidence that can be checked against global ultrasound appearance and diagnostic context.

\begin{table}[h]
\centering
\caption{\textbf{Detector-agnostic renal diagnosis results.} Replace our detector with different detectors and evaluate overall accuracy (\%) on the renal validation split. The compared detectors include YOLO-family models~\citep{yolo,yolov11,yolov12,yolo26}, LW-DETR-DINOv3~\citep{lw-detr,dinov3}, and RF-DETR-DINOv3~\citep{rf-detr,dinov3}. Ours denotes our detector. The right panel visualizes the same detector-only and agent-with-detector accuracies.}
\label{tab:detector_ablation}
\setlength{\tabcolsep}{3pt}
\renewcommand{\arraystretch}{1.05}
\begin{tabular}{@{}m{0.43\linewidth}m{0.53\linewidth}@{}}
\begin{minipage}[c]{\linewidth}
\centering
\vspace*{2pt}
{\small
\begin{tabular}{@{}lrrr@{}}
\toprule
Detector & Det. & \echoname{} & Gain\\
\midrule
YOLOv8 & 44.16 & 67.11 & +22.95\\
LWDETR-DINOv3 & 45.91 & 56.08 & +10.17\\
YOLO26 & 64.56 & 71.01 & +6.45\\
RFDETR-DINOv3 & 65.91 & 69.70 & +3.79\\
YOLO12 & 66.85 & 72.35 & +5.50\\
YOLO11 & 68.99 & 71.14 & +2.15\\
Ours & \textbf{74.53} & \textbf{77.43} & +2.90\\
\bottomrule
\end{tabular}
}
\end{minipage}
&
\begin{minipage}[c]{\linewidth}
\centering
\includegraphics[width=0.94\linewidth]{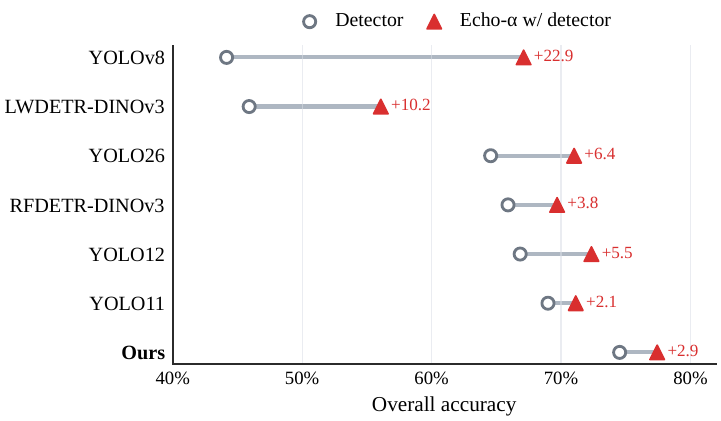}
\end{minipage}
\end{tabular}
\end{table}

The gain differences reveal a consistent rule: \echoname{} improves weaker detectors more, but the source of improvement depends on the detector's candidate structure.
For sparse and selective detectors, the main residual error is missed positive evidence; YOLOv8, for example, covers only 31.4\% of images with nearly one high-confidence candidate per covered image, so \echoname{} uses reliable local anchors to recover missing diagnostic context and obtains the largest gain (+22.95 points).
For dense detectors, the residual error shifts from absence of evidence to ambiguity among competing candidates; LWDETR-DINOv3 covers 95.2\% of images with 2.16 candidates per covered image on average, so \echoname{} mainly improves diagnosis by verifying and re-ranking noisy evidence (+10.17 points).
As detector accuracy increases, both types leave less correctable error, which explains the smaller but still positive gains on stronger backbones.
Thus, these results suggest that \echoname{} can adapt its correction behavior to the detector's error mode: contextual completion for sparse outputs and evidence suppression for dense outputs.

\subsection{Case Study}
\begin{figure}[t]
\centering
\includegraphics[width=\linewidth]{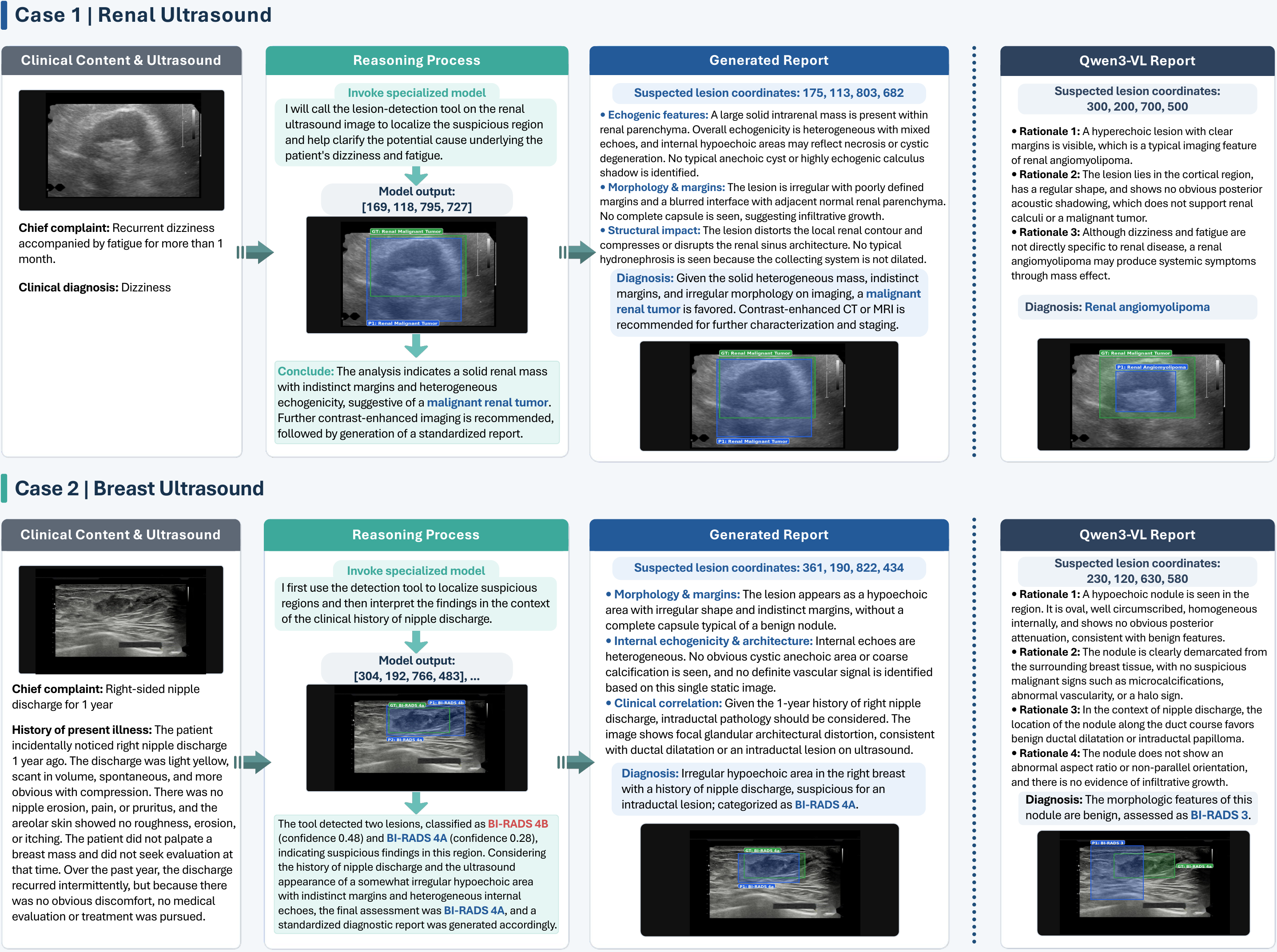}
\caption{\textbf{Case studies of \echoname.} In both renal and breast ultrasound, the detector provides candidate evidence, the grounding-oriented setting improves spatial alignment to the true lesion, and the diagnosis-oriented setting further revises the final prediction by integrating tool feedback with lesion appearance, surrounding context, and clinical cues. Across both cases, \echoname{} uses tool-grounded evidence to distinguish among candidate interpretations, rather than simply reproducing either the raw report or the detector output.}
\label{fig:case}
\end{figure}

Figure~\ref{fig:case} gives two representative examples and shows that the further value of \echoname{} is the better evidence alignment.
\textbf{Renal ultrasound.} The raw Qwen3-VL~\citep{Qwen3-VL} report favors benign angiomyolipoma from a smaller benign-looking region, while \echoname{} grounds a larger lesion and predicts malignant renal tumor.
The difference is consistent with detector-grounded evidence of heterogeneous echoes, indistinct margins, and local structural distortion.
\textbf{Breast ultrasound.} The raw Qwen3-VL report gives a benign BI-RADS 3 assessment, while \echoname{} predicts BI-RADS 4A after the tool returns two ductal-region candidates, BI-RADS 4B and BI-RADS 4A.
Rather than following the higher-confidence 4B candidate, \echoname{} selects the lower-confidence 4A candidate by integrating the grounded irregular hypoechoic pattern with optional clinical information such as a 1-year history of nipple discharge.

Across both cases, \echoname{} uses tool-grounded evidence to distinguish among candidate interpretations, rather than simply reproducing either the raw report or the detector output.

\section{Conclusion}
We introduced \echoname, an agentic multimodal framework that unifies detector-level precision with MLLM-based reasoning for ultrasound interpretation.
Its central idea is simple but effective: rather than treating detector outputs as final predictions, \echoname{} treats them as callable evidence inside a learnable invoke-and-reason loop.
Starting from a shared supervised model and applying different reward optimization toward grounding-oriented and diagnosis-oriented objectives yields two complementary behaviors, \echoground{} and \echodiag, that together improve both lesion localization and final clinical decision-making under a multi-center evaluation protocol.
Beyond the empirical gains, our results suggest a broader lesson for medical AI: reliable multimodal reasoning is more likely to emerge when large models are trained to interrogate specialized tools, verify their outputs, and integrate them with global clinical context.
We hope \echoname{} provides a practical foundation for building interpretable, extensible, and clinically meaningful agentic systems across ultrasound and other medical imaging domains.

\end{document}